\documentclass[10pt,twocolumn,letterpaper]{article}

\usepackage{cvpr}              
\usepackage{float}
\definecolor{cvprblue}{rgb}{0.21,0.49,0.74}
\usepackage[pagebackref,breaklinks,colorlinks,allcolors=cvprblue]{hyperref}

\def\paperID{19} 
\def\confName{CVPR}
\def\confYear{2026}

\title{From Synthetic Data to Real Restorations: Diffusion Model for Patient-specific Dental Crown Completion}

\author{
Dávid Pukanec \quad Tibor Kubík \quad Michal Španěl\\
Department of Computer Graphics and Multimedia,\\
Brno University of Technology, Czechia\\
{\tt\small \{ipukanec, ikubik, spanel\}@fit.vutbr.cz}
}

\begin{document}
\twocolumn[{%
\renewcommand\twocolumn[1][]{#1}%
\maketitle
\begin{center}
    \includegraphics[width=\textwidth]{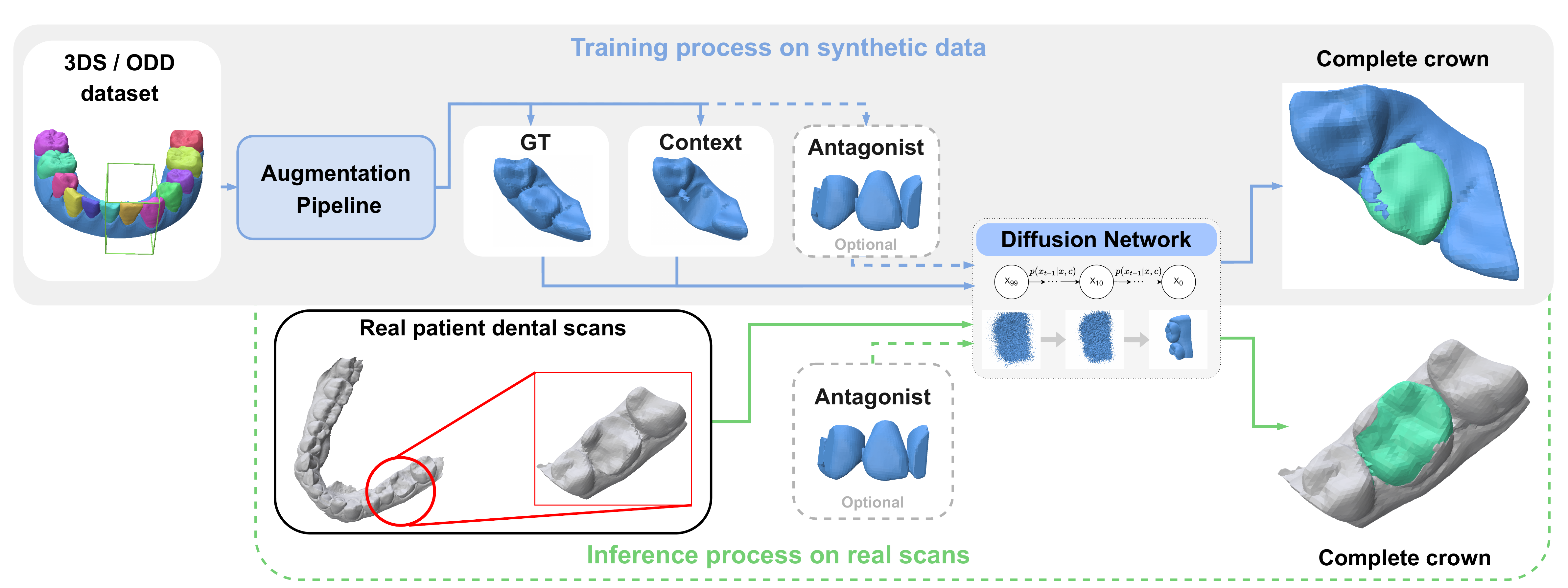}
    \captionof{figure}{High-level overview of tooth completion process. A unified model is conditioned on the local context, with the optional antagonist condition. The incomplete shapes are synthetically generated. Our model works with every tooth class. The input data are represented as signed distance fields.}
    \label{fig:teaser}
\end{center}%
}]
\begin{abstract}

We present \textbf{ToothCraft}, a diffusion-based model for the contextual generation of tooth crowns, trained on artificially created incomplete teeth. Building upon recent advancements in conditioned diffusion models for 3D shapes, we developed a model capable of an automated tooth crown completion conditioned on local anatomical context. To address the lack of training data for this task, we designed an augmentation pipeline that generates incomplete tooth geometries from a publicly available dataset of complete dental arches (3DS, ODD). By synthesising a diverse set of training examples, our approach enables robust learning across a wide spectrum of tooth defects. Experimental results demonstrate the strong capability of our model to reconstruct complete tooth crowns, achieving an intersection over union (IoU) of 81.8\% and a Chamfer Distance (CD) of 0.00034 on synthetically damaged testing restorations. Our experiments demonstrate that the model can be applied directly to real-world cases, effectively filling in incomplete teeth, while generated crowns show minimal intersection with the opposing dentition, thus reducing the risk of occlusal interference. Access to the code, model weights, and dataset information will be available at:~\url{https://github.com/ikarus1211/VISAPP_ToothCraft}.
\end{abstract}    
\section{Introduction}
\label{sec:introduction}

Restoration of a damaged or missing tooth is a procedure that aims to reconstruct its original shape. Typically, a technician selects a predefined tooth from a template and sculpts it to meet both functional and aesthetic requirements. The introduction of CAD/CAM systems has greatly enhanced this process and provided the ability for same-day tooth fabrication~\cite{muhlemann2021production}. However, the design of the restoration remains very time-consuming, requires specialised skills, and is prone to human error. Therefore, there is a growing interest in automating this process \cite{fruehwirt2021towards}. In recent years, various studies have been conducted to tackle a related but broader issue known as shape completion~\cite{yan2022shapeformer,Mittal_2022_CVPR,wu2020multimodal,chu2023diffcomplete}. This motivates us to explore deep learning methods to address the problem of tooth crown restoration.

Denoising Diffusion Probabilistic Models (DDPMs) \cite{ho2020denoising} have emerged as one of the most promising classes of generative models in recent years. These models have shown remarkable performance across various 2D tasks \cite{rombach2022high,podell2023sdxl}, as well as in the area of 3D shape generation \cite{luo2021diffusion}. To steer the generative process toward the desired output, these diffusion models can be conditioned using auxiliary information, such as partial input, class labels, or text~\cite{zhang2023adding,ho2022classifier,rombach2022high}. This inspires the idea of using diffusion models for tooth reconstruction, conditioned on the local context composed of neighbouring teeth and opposing dentition. 

However, a key challenge is that diffusion models require large datasets~\cite{rombach2022high,NEURIPS2024_f782860c}, raising the question of how such extensive data can be obtained. Acquiring datasets in dentistry can be problematic due to high costs, privacy concerns, and ethical liabilities. Simultaneously, diffusion-based approaches often require thousands of samples for learning. Nevertheless, only a limited number of publicly available 3D shape dental datasets exist~\cite{ben2022teeth3ds,wang20243d}. Furthermore, none of these datasets specifically includes incomplete or defective teeth for the purpose of tooth completion. To address this issue, we devised an augmentation pipeline that modifies healthy teeth from publicly available datasets of segmented dental arches, allowing us to train a single, unified diffusion model that can, in a self-supervised fashion, restore a wide variety of tooth defects or missing teeth, as depicted on Figure~\ref{fig:teaser}.

Our core contributions are summarised as follows:
\begin{itemize}
    \item  We are the first to present a diffusion-based architecture, employing the local anatomical context as a condition to create a single unified model capable of restoring various tooth defects and generating whole teeth from digital dental casts.
    
    \item We have designed an augmentation pipeline that generates incomplete tooth samples from datasets of complete dental arches. This enables training a conditional diffusion model on synthetic data in a self-supervised fashion for the tooth completion task.
    
    \item The evaluation shows that the model trained on synthesised incomplete samples can effectively transfer the learned knowledge to real-world scans of incomplete tooth data.

\end{itemize}

\section{Related work}
\label{sec:related_work}
\textbf{3D Shape Completion}. The completion task from the deep learning perspective involves inferring a complete and plausible geometry of an object or scene from an incomplete or degraded shape. Since 3D shapes can be represented in various ways, a range of approaches have been developed to address these different representations. The Point Completion Network (PCN)~\cite{yuan2018pcn} introduces a coarse-to-fine decoder that first predicts a global shape outline and then refines it. SnowflakeNet~\cite{Xiang_2021_ICCV} uses a unique "Snowflake Point Deconvolution". GRNet~\cite{xie2020grnet} uses a grid structure to regularise point clouds. 3D-EPN~\cite{dai2017shape} uses an encoder-decoder architecture with a low-resolution voxel, refined using shape priors. Generative models~\cite{Mittal_2022_CVPR,wu2020multimodal} have shown promising results, although their training typically requires large datasets. The emergence of Transformers~\cite{vaswani2017attention} has led to the development of ShapeFormer~\cite{yan2022shapeformer}. The introduction of attention significantly improved the capture of long-range dependencies and global structure. However, the high memory cost associated with attention often makes these approaches infeasible for large resolution grids. A notable diffusion-based approach is DiffComplete~\cite{chu2023diffcomplete}, which employs a voxel-based representation and innovative occupancy-aware fusion to achieve the latest results. The method aims to complete a model by merging multiple range scans, which can capture an arbitrary portion of the shape. 

\smallskip
\noindent
\textbf{Tooth Crown Generation.} Several works have been proposed in the field of tooth crown generation. DCrownformer~\cite{yang2024dcrownformer} introduces a novel transformer-based architecture to generate dental meshes from point clouds. MVDC~\cite{yang2025mvdc} propose a contrastive learning network trained on depth map projections of 3D shapes. The study focuses only on the first mandibular molars and is limited to partial tooth damage. \cite{hosseinimanesh2025personalized} proposes a transformer-based architecture utilising incomplete point clouds for personalised crown modelling. They used a partial local context to design a crown fabricated from a prepared tooth stump. Although similar to our task, this approach focuses solely on the crown design from the stump and does not address cases where whole teeth are missing or partial fillings (inlays or onlays) are required. \cite{ji20253d} introduces the diffusion model for the reconstruction of missing teeth with a separate module for the prediction of the position of the teeth. Although the authors do not address the full problem of tooth completion, they primarily concentrate on predicting accurate tooth positions and overall shapes based on geometries reconstructed from CBCT images, which themselves exhibit limited data diversity. \cite{chafi2025exploring} proposes a method based on GANs to generate preparation stumps conditioned on specifically designed crowns. This approach allows for greater accuracy compared to the commonly used sphere prior. However, this work did not focus on generating complete tooth crowns. \cite{kubik2025toothforge} explored deep spectral representation for crown population modelling, which, however, allows only generation that is not conditioned on the context.

In summary, existing shape completion methods are successful in 3D object completion, but they lack dental specificity and fail to address all needs in tooth completion. The tooth crown completion methods discussed focus primarily on specific tooth types, partial defects, or crown design from prepared stumps. In this work, we address the tooth completion problem from the perspective of general shape completion and design an architecture aimed at solving this issue.
\section{Method}
Our goal was to develop ToothCraft, a diffusion-based generative model capable of producing anatomically accurate teeth from incomplete or missing tooth structures, while considering the local anatomical context. As shown in previous work~\cite{hosseinimanesh2025personalized}, the context offers valuable insights into the morphology, style, and occlusal relationship of the tooth. To guide our model, we draw on key concepts from ControlNet \cite{zhang2023adding} and DiffComplete \cite{chu2023diffcomplete}, both of which demonstrate strong controllability. Additionally, our objective is to train our network using synthesised incomplete tooth data derived from publicly available datasets~\cite{ben2022teeth3ds,wang20243d} of segmented dental arches, while ensuring that the model performs well on real cases.

\subsection{Teeth Completion Architecture}
Diffusion models consist of two processes, the \textit{forward} noising process and \textit{backward} denoising process. The forward process $q(x_{1:T}|x_0)$ progressively adds Gaussian noise to the input $x_0$ to transform it into a pure noise distribution. Here we use the number of timesteps $T=1000$. The \textit{backward} process $p(x_{0:T}, c)$ reconstructs a clean sample, conditioned on $c$ by iteratively predicting the mean sample $\mu_\theta(x_t,t,c)$ in a timestep $t \in \{1,..., T\}$. In the case of ToothCraft, a more common and computationally straightforward formulation is used, in which the model is trained to predict the added noise $\epsilon_\theta(x_t, t, c)$ while minimising the standard diffusion loss:
\begin{equation}
        \arg\min E_{t,x_o,\epsilon,c}(|| \epsilon - \epsilon_\theta(x_t,t,c)||^2), ~ \epsilon \in \mathcal{N} (0,I),
\end{equation}

where $\epsilon$ represents the noise applied to the clean input to create $x_t $ in the \textit{ forward process}. To achieve better control over the diffusion process and increase the diversity of samples, the method can be extended to classifier-free guidance~\cite{ho2022classifier}. During training, the model learns to predict noise both with and without conditioning,  where the conditioning signal is randomly omitted with a probability of 0.15. At inference time, the predicted noise is formulated as:
\begin{equation}
    \hat{\epsilon}_\theta = \epsilon_\theta(x_t) + w(\epsilon_\theta(x_t|c) - \epsilon_\theta(x_t))
\end{equation}
where $\hat{\epsilon}_\theta$ represents the noise predicted by classifier-free diffusion and $w$ denotes the mixing factor. 

\begin{figure*}
    \includegraphics[width=\linewidth]{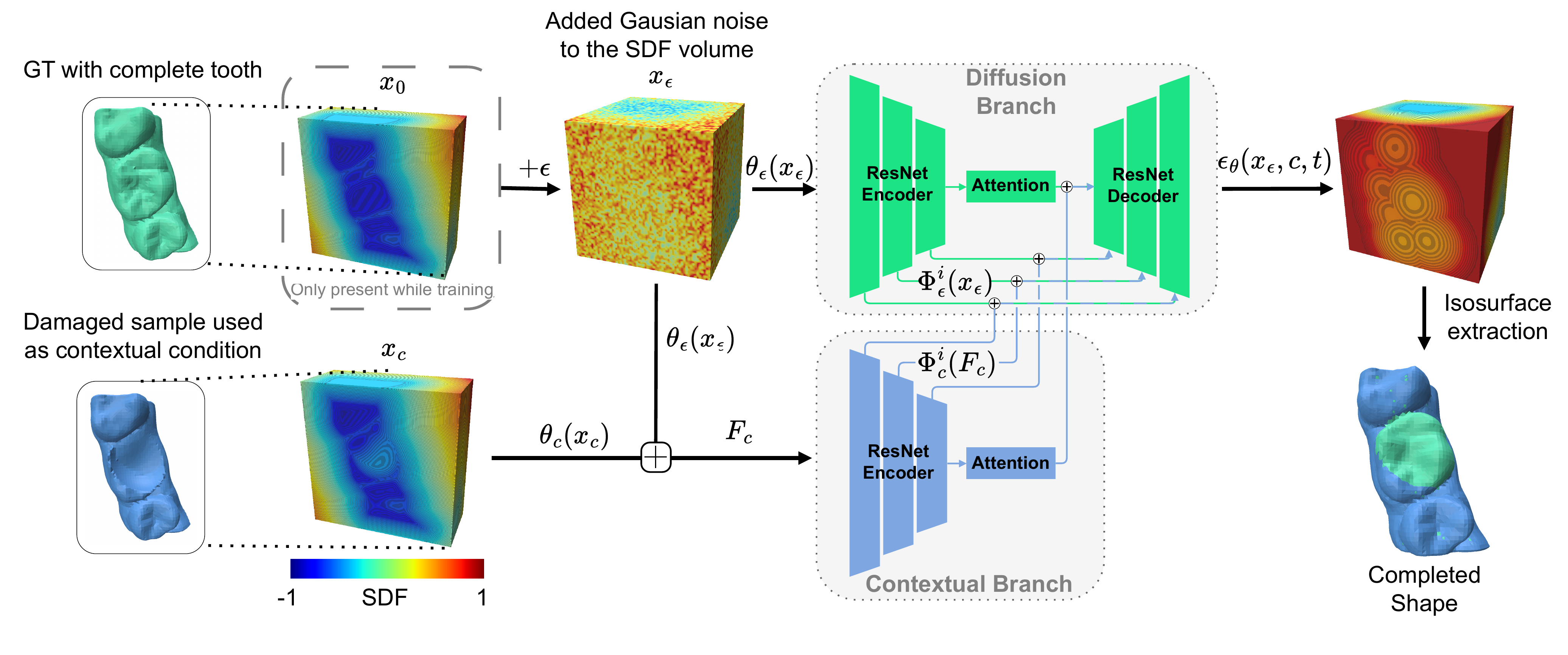}
    \caption{Detailed overview of completion architecture. For the additional antagonist, the Contextual branch is duplicated, and feature vectors are added together. Symbol \includegraphics[width=0.020\textwidth]{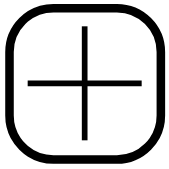} represents concatenation.}
    \label{fig:arch}
\end{figure*} 

The input of ToothCraft consists of the local anatomical context of the tooth to be completed. This input is generated using our pipeline, as explained in Section~\ref{sec:augpip}, which is represented by a Signed Distance Field (SDF).
The method can be divided into several branches based on the input.

\textbf{Diffusion Branch.} ToothCraft is built on the aforementioned diffusion formulation using a UNet backbone~\cite{rombach2022high}, extended to 3D, that performs denoising, denoted as the Diffusion branch, see Figure~\ref{fig:arch}. The branch takes a complete shape and adds noise to it. This noisy SDF volume $x_\epsilon$ that is projected through two $1\times1$ convolutions $\theta_\epsilon(x_\epsilon)$ to a higher dimensionality and encoded by ${\Phi_{\epsilon}(\cdot)}$. This encoder progressively downsamples the volume and only uses attention in the middle block to reduce computational cost. 

\textbf{Contextual branch.} Following the architecture of the Diffusion branch, the Contextual branch takes an input, the incomplete context $x_c$, represented as an SDF. Similarly, $x_c$ is fed through projection convolution $\theta_c(x_c)$. The resulting context features are then concatenated with the features of the noisy input $F_c = [ \theta_\epsilon(x_\epsilon),\theta_c(x_c)]$, where $[\cdot,\cdot]$ is concatenation. This fused feature volume is then passed through the context encoder ${\Phi_c(F_c)}$. The decoder $\Psi_{\epsilon}(\cdot)$ is shared between both the Diffusion and Context branches. For each decoder layer $i$, the decoder features can be expressed as:
\begin{equation}
d^i_{xc} = [\Psi^{i-1}_{\epsilon}(x_\epsilon),\Phi^i_{\epsilon}(x_\epsilon) + \Phi^i_c(F_c)].
\end{equation} This per-level feature aggregation enables the network to condition the output in multiple levels.

\textbf{Antagonist Conditioning.} In tooth restoration, opposing teeth play a crucial role in understanding the morphological structure of the restored tooth due to their occlusal interactions. As this antagonist information might not always be available, the decision was made to split it into separate inputs. To facilitate conditioning on the antagonist, a third encoder $\Phi^i_a$ is added. It shares the same architecture as the contextual one. Its features are combined using the addition as:
\begin{equation}
d^i_{xca} = [\Psi^{i-1}_{\epsilon}(x_\epsilon), \Phi^i_{\epsilon}(x_\epsilon) + \Phi^i_c(F_c) + \Phi^i_a(F_a)],
\end{equation}
where $F_a$ represents the input features of the antagonist encoder $\Phi^i_a$ computed as $F_a = [ \theta_\epsilon(x_\epsilon),\theta_a(x_a)] $. Note that the antagonist and context encoders do not share weights. This is intentional as they serve different purposes: the context encoder guides the diffusion toward the presence of shape, while the antagonist encoder steers it away from regions occupied by the opposing tooth.

\subsection{Break Teeth to Fix Teeth}\label{sec:augpip}

To train the method mentioned above, we require incomplete teeth along with their corresponding ground truth. As of now, we are not aware of any existing publicly available dataset including such samples. Consequently, we generate our own training samples through an augmentation pipeline, using the publicly available dental arch datasets Teeth3DS~\cite{ben2022teeth3ds} and the Orthodontic Dental Dataset (ODD)~\cite{wang20243d}, both of which provide surface meshes of segmented dental arches. The goal of the pipeline is not to simulate or replicate realistic anatomical defects, as this would be unrealistic. Instead, the goal is to create a diverse set of samples that exhibit varying degrees of incompleteness. This approach, similar to techniques used with images \cite{lugmayr2022repaint}, is designed to encourage the model to develop contextual reasoning skills by introducing a certain level of incompleteness. Training on these diverse partial inputs should yield a robust latent diffusion distribution applicable to many real-world scenarios, eliminating the need for datasets with real scans and actual crown damage cases.
\begin{figure*}
    \centering
    \includegraphics[width=\linewidth]{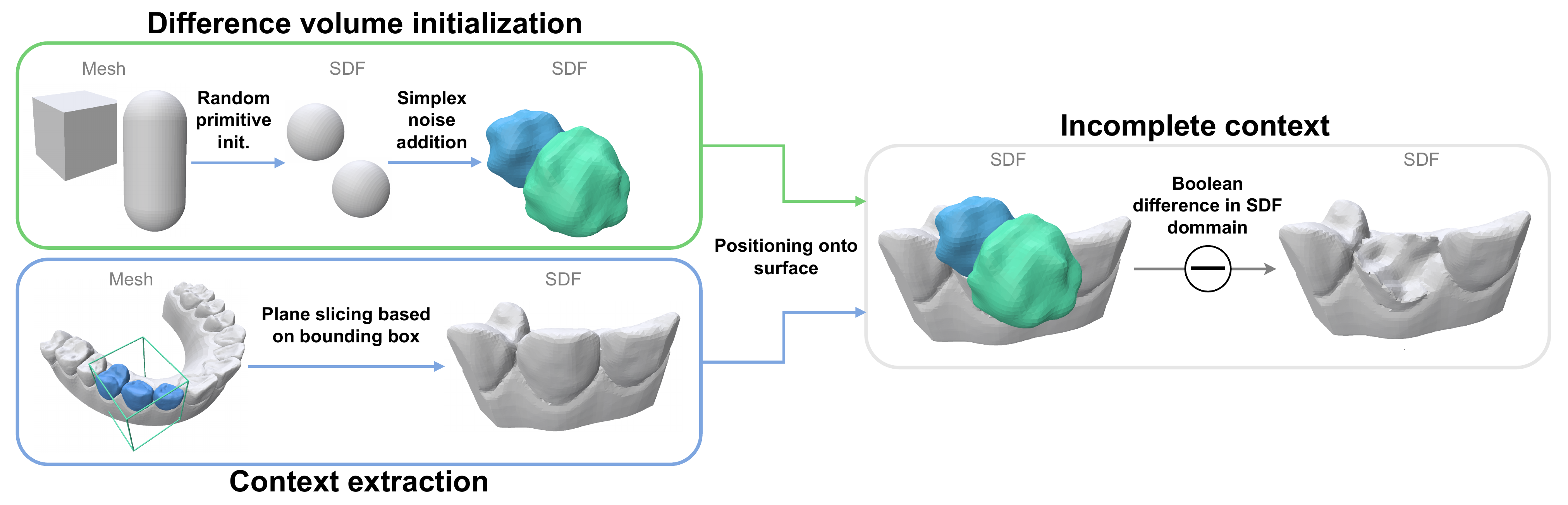}
    \caption{Augmentation pipeline. Internal representation is denoted above the visualised meshes.}
    \label{fig:augpip}
\end{figure*}

The pipeline takes a segmented dental arch mesh as input. The first step is the extraction of the tooth together with the surrounding context, which is then normalised and converted to an SDF using the method of~\cite{wang2022dual}, as depicted in Figure~\ref{fig:augpip}. This unmodified tooth serves as the ground truth. To generate an incomplete tooth, we sample random primitives from a predefined set (\textit{sphere, cube, cylinder, capsule, cone}), randomly transform them, and centre them on the surface of the tooth. The subsequent trim of the primitive with the bounding box of the tooth avoids excessive changes to neighbouring teeth during Boolean operations. Next, we convert these shapes into an SDF representation and apply simplex noise $s(\cdot)$ to  voxels with a center at $\mathbf{v}$ of SDF $x_{\mathrm{sdf}}(\mathbf{v})$, producing perturbed primitive $ \tilde{x}_{\mathrm{sdf}}(\mathbf{v})$ as follows:
\begin{equation}
\tilde{x}_{\mathrm{sdf}}(\mathbf{v}) = x_{\mathrm{sdf}}(\mathbf{v}) + \alpha\, s(f\mathbf{v}), 
\end{equation}
where $\alpha= 0.06$ is the amplitude and $f = 2.8$ is the frequency. This addition disturbs the smooth surface and adds more complexity to the primitives. The addition of the simplex noise is crucial, as without it, the network tended to complete shapes only where smooth surfaces existed, which hindered its ability to generalise well on real samples. The final step involves performing a Boolean difference between the tooth and the primitives in the SDF representation. In this manner, we obtain paired incomplete samples and their corresponding ground truth. Moreover, this type of augmentation allows us to reuse the same contextual region multiple times with varying patterns of synthetically applied damage, thereby improving the data diversity and robustness of the model.
\section{Experiments}
\label{sec:Experiments}
\label{sec:model_val}
\begin{table*}[!h]
  \centering
    \caption{Results reported over the generated test set. We report metrics calculated over the whole SDF volume and also the masked volume. Bold numbers represent the best results of the mean across the entire testing set. For clarity, CD values were multiplied by $10^4$.}
  \begin{tabular}{c|@{\hspace{4pt}}cc@{\hspace{6pt}}cc}
    \hline
    Model &   L1  & mL1 & CD  & mCD   \\
    \hline
    Normal  &  $\mathbf{0.0169 \pm 0.0281}$  &  $0.0316 \pm 0.0245$  
    &  $4.841 \pm 11.682$  &  $ 3.122 \pm 7.329$  \\
     
    Classifier Free    &  $0.0184 \pm 0.0311$  &  $\mathbf{0.0266 \pm 0.0156}$  
    &  $\mathbf{3.223 \pm 4.372}$  &  $2.289 \pm 3.561$ \\
     
    Antag &  $ 0.0207 \pm 0.0361$  &  $0.0281 \pm 0.0216$ 
    &  $ 3.427 \pm 5.9$   &  $\mathbf{2.041 \pm 2.717}$  \\
    
    \hline
    Model &  IoU $\uparrow$ & mIoU $\uparrow$ & IoU $\mathrm{Antag}_{pred}$ $\downarrow$ &  IoU $\mathrm{Antag}_{gt}$ $\downarrow$  \\
    \hline
    
    Normal  &  $79.54\%  \pm 13.13\% $  &  $84.63\%  \pm 10.15\% $  
     &  $0.38\% \pm 0.74\%$  &  $0.07\%  \pm 0.2\% $   \\
     
     Classifier Free    &  $81.5\% \pm 10.8\%$  &  $\mathbf{87.3\% \pm 6.7\%}$  
     &  $0.2\% \pm 0.4\%$  &  $0.07\% \pm 0.2\% $ \\
     
    Antag &  $\mathbf{ 81.8\%  \pm 11.5\%  }$  &  $85.4\%  \pm 9.9\% $ 
    &   $\mathbf{ 0.1\%  \pm 0.3\% }$   &  $0.07\% \pm 0.2\% $  \\
    \hline
\end{tabular}

  \label{tab:generated}
\end{table*}

While the process of generating synthetic data is well-suited for training, we recognise that evaluating only such data would not adequately reflect real clinical use. Therefore, in addition to tests on synthetic samples, we also conducted an evaluation on 16 real-world cases of tooth defects, with ground truth created by clinicians. These examples were provided by the TESCAN company and show various crown defects, ranging from partial tooth loss to complete tooth absence. Examples also include gingival abnormalities, such as screws and abutments for stump and crown fixation, and gingival recessions. Using these examples, we tested the network's ability to transfer knowledge from synthetic damage to real cases and its potential for clinical application.

\subsection{Evaluation Metrics}

We evaluate accuracy using metrics computed between the completed SDF and the ground truth: the average distance per-voxel $L1$, the bidirectional Chamfer distance $CD$ derived from the extracted meshes, and the intersection based on the voxel over union $IoU$. To focus specifically on the completion volume, we propose \textit{masked} evaluation metrics which are calculated only within the axis-aligned bounding box of the ground truth tooth. These masked metrics are denoted as $mL1$, $mCD$, and $mIoU$.  These metrics provide a more reliable assessment of the accuracy of the completion, as the boundary box encompasses the tooth, which corresponds to the region of clinical relevance when designing prostheses. To evaluate the interference of the occlusal surface between the completed tooth and its antagonist, we report IoU $Antag_{pred}$, which measures the percentage of voxels where the predicted completed tooth intersects with its antagonist, relative to the union of these voxels. A similar metric is the IoU $Antag_{gt}$, which indicates the original interference present in the dataset between the ground truth sample and its antagonist. 

\subsection{Training and Data}
\label{sec:train_data}

\begin{figure*}
    \centering
    \includegraphics[width=0.90\linewidth]{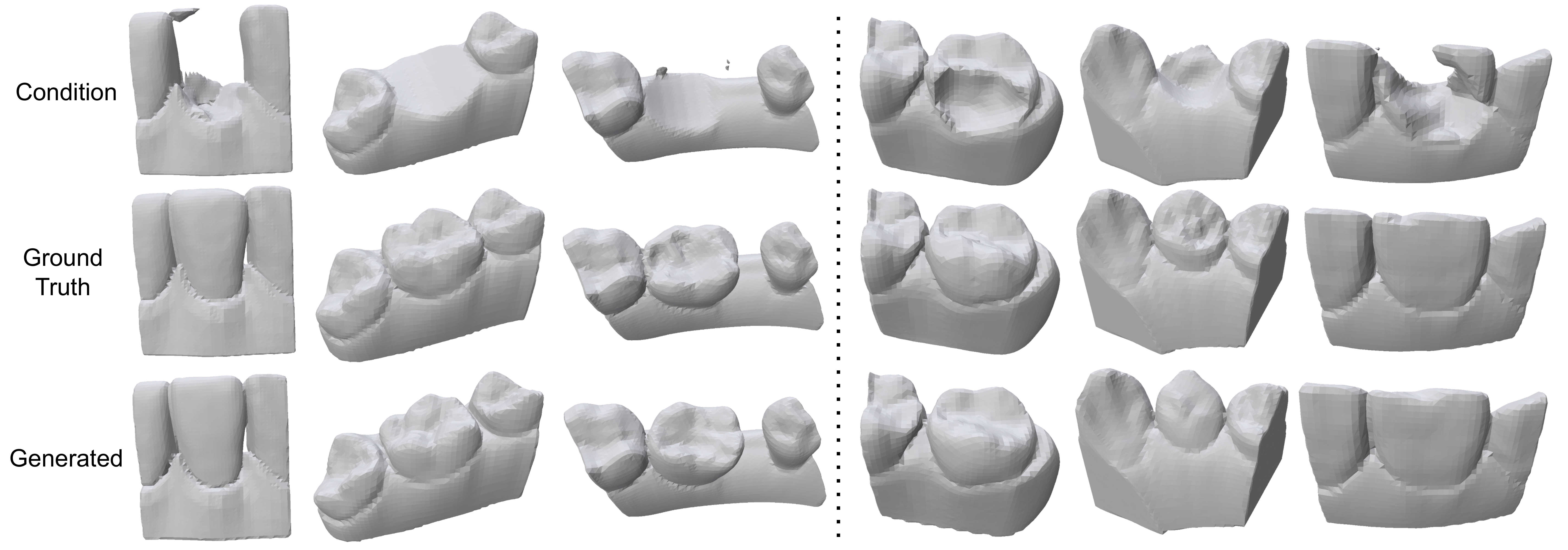}
    \caption{Visual representations of meshes obtained through the Marching Cubes algorithm on test set.}
    \label{fig:test_res}
\end{figure*}

The network was trained for 700k iterations with a batch size of 8, with a grid resolution of 64. Training was carried out using 1 $\times$ Nvidia H100 NVL, the peak memory allocation with this batch size was 60 GB. It took 110 hours to train the model on a single GPU. The learning rate was fixed at $1 \times 10^{-4}$. For sampling the diffusion timesteps $t$, we employed a loss-aware second-moment resampler, which emphasises learning the timesteps associated with higher prediction errors. 

The training data set was produced before training using the pipeline mentioned above in Section \ref{sec:augpip} with the ODD data, where 810 randomly sampled training arches were used, resulting in 20568 generated samples of local context. The number of primitives capable of damaging the tooth was limited to a maximum of three, each with a relative size between 0.2 and 0.5 of the modified tooth size. Each arch was used only once in the context generating pipeline, meaning each local context is distinct. The exact data splits for the train, validation and test sets will be provided with the code.

\subsection{Model Validation}
\label{sec:modval}

In this section, we present the evaluation results of our generated data on the test set. This evaluation aims to assess how effectively the model has learned to perform the general crown completion. To conduct this assessment, we utilised a test set generated by our pipeline from dental arches that were not part of the training data creation process. This approach ensures that the generated teeth exhibit unique morphological contexts that have not been previously encountered. The testing set consisted of more than 5398 samples generated from 224 arches. We performed the evaluation using timestep respacing~\cite{nichol2021improved} with 100 steps, to speed up the diffusion process.

To conduct a more thorough evaluation, we trained three distinct models: (1) \textit{Normal} model trained without the antagonist information, (2) \textit{Antag} model trained with antagonist information and (3) \textit{Classifier Free} model trained with a condition dropout of 10\% and mixing factor ($w$) was set to $2.0$. The results are presented in Table~\ref{tab:generated} with a visual representation in Figure~\ref{fig:test_res}. The results indicate that the \textit{Normal} model achieves a very low L1 distance. However, the high standard deviation suggests that the model generates more unstable samples. In contrast, the \textit{Classifier Free} model exhibits a relatively low standard deviation across nearly all metrics. This supports the claim that training a model using the classifier-free approach leads to more stable diffusion. The \textit{Antag} model, while having the highest \(L1\) distance, demonstrates a greater ability to infer the overall shape as the IoU and CD metrics are substantially better than the \textit{Normal model}. Moreover, it produced a very low intersection between the generated samples, nearly aligning with the average mean of the generated dataset. This emphasises the significance of conditioning the model on the antagonists to reduce occlusal interference.

\subsection{From Synthetic to Real Data}
\begin{figure*}
    \centering
    \includegraphics[width=\linewidth]{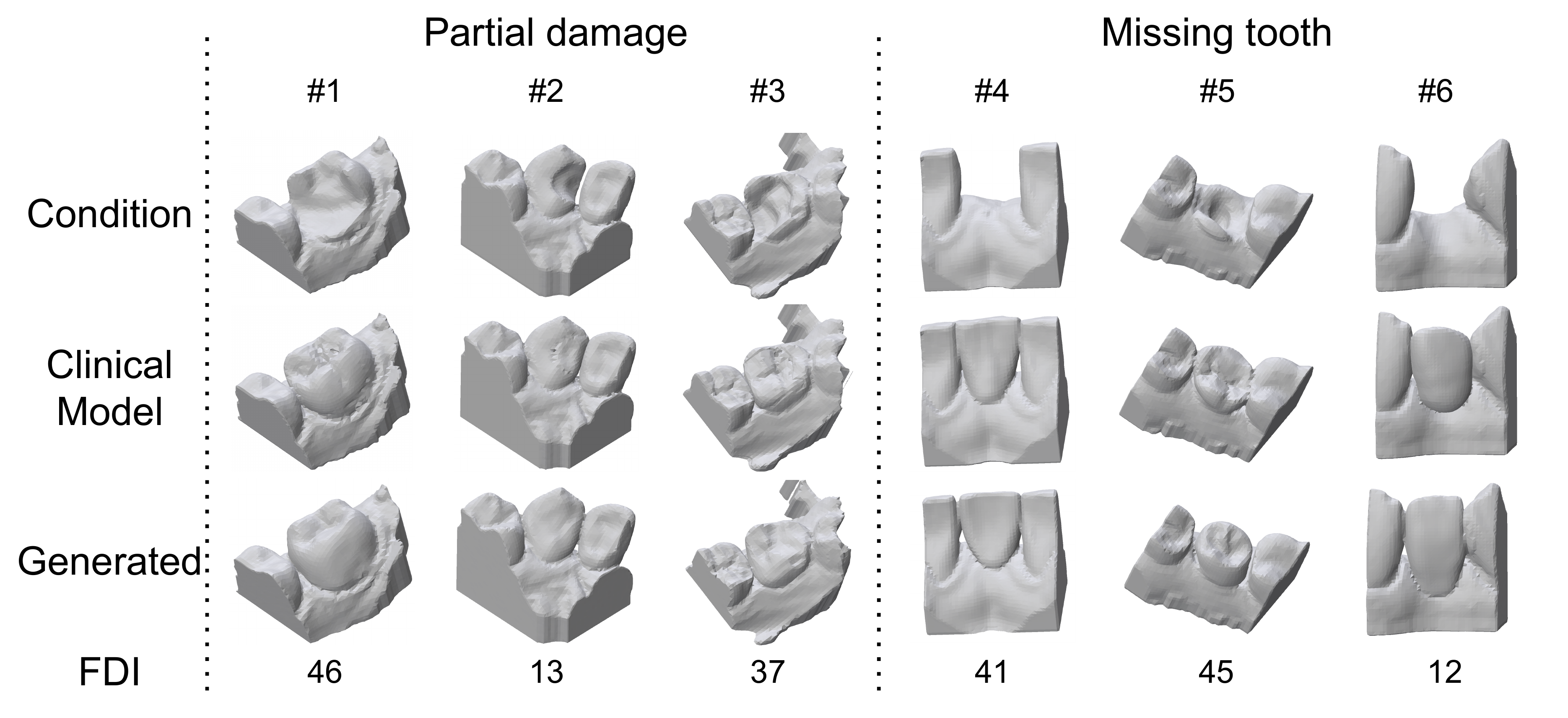}
    \caption{Examples of teeth completed by our network. All results were achieved using a single cohesive model, with human-designed restorations crafted in laboratories by technicians.}
    \label{fig:real_example}
\end{figure*}
\begin{figure}
    \centering
    \includegraphics[width=\linewidth]{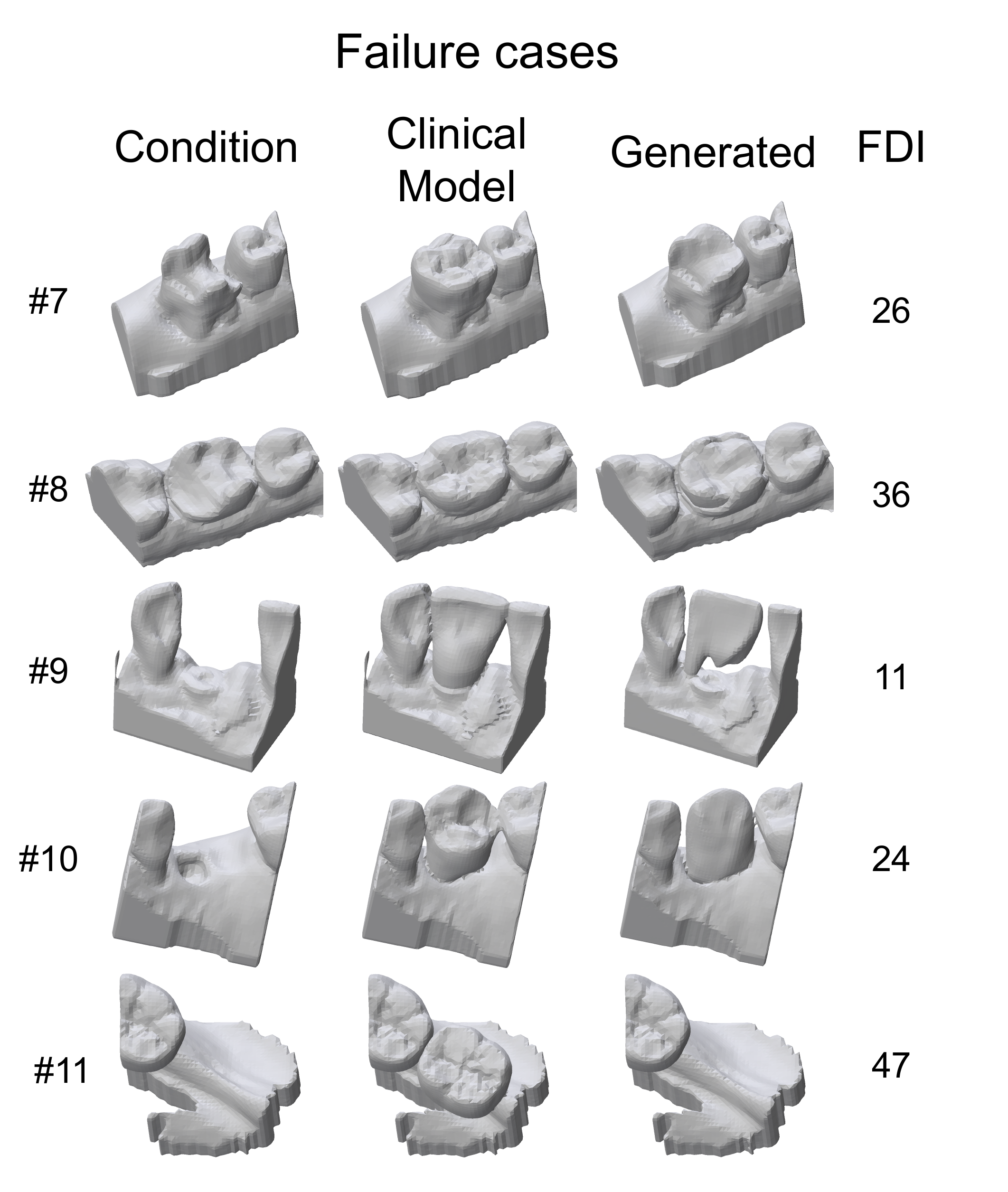}
    \caption{Failure cases produced by our model. The most common reason for failing to complete a model is gingiva abnormality.}
    \label{fig:failure}
\end{figure}
\begin{table}[]
    \centering
    \caption{The metrics computed between generated samples and complete models provided by technicians. We only report the masked version as it gives a more accurate result regarding the completed region. For clarity, CD values were multiplied by $10^4$.}
    
\begin{tabular}{cc|cccc}

    Case & FDI &  mL1 & mIoU & mCD \\
    \hline
    \#1 & 46 &0.0820&  46.8\% & 4.74 \\
   
    \#2 & 13 &0.0378 & 60.5\% & 1.88 \\
    
    \#3 & 37 &0.0366 & 70.3\% & 2.46 \\ 
    
    \#4 & 41 &0.0208 & 83.7\% & 0.95 \\
    
    \#5 & 45 &0.0281 & 80.4\% & 2.54 \\
    
    \#6 & 12 &0.0614 & 61.1\% & 4.17 \\ 
    
    \#7 & 26 & 0.0413 & 67.2\% & 2.39 \\ 
    
    \#8 & 36 & 0.0590 & 53.4\% & 5.84 \\
    
    \#9 & 11 & 0.0327 & 81.9\% & 1.9 \\ 
    
    \#10 & 24 & 0.0530 & 64.1\% & 4.78 \\
    
    \#11 & 47 & 0.1950 & 15.8\% & 5.57 \\
    
    \#12 & 25 & 0.0327 & 81.9\% & 1.94 \\
    
    \#13 & 13 & 0.0784 & 50.9\% & 5.57 \\
    
    \#14 & 32 & 0.0518 & 68.2\% & 3.63 \\
    
    \#15 & 46 & 0.0765 & 43.0\% & 5.81 \\ 
    
    \#16 & 36 & 0.0297 & 73.1\% & 1.31 \\  
\end{tabular}

    \label{tab:real_metrics}
\end{table}

Now that we have demonstrated that the model can complete the tooth using synthetic data, we will evaluate its performance on real scans with various defects. These samples were completed using the same model and setup as the results categorised as \textit{Normal} in~~ Section~\ref{sec:model_val}. We manually cut out the region with the surrounding context, converted it to an SDF of resolution 64, and fed it to the network. For these samples, we do not have an antagonist arch, but we were provided with a modelled complete tooth created by technicians. In addition to providing a visual comparison in Figures~\ref{fig:real_example}  and~\ref{fig:failure}, we calculated the metrics as previously, visible in Table~\ref{tab:real_metrics}. It is important to note that in this case, poor metrics do not necessarily indicate that the network failed to generate a plausible tooth. If the same task were given to ten technicians, each would model the tooth in different ways, leading to similarly inconsistent metrics. Our goal with the visual representation is to show how well the overall morphological structure fits the context, how well the tooth is aligned with the neighbouring teeth, whether it successfully models the interdental space, and the level of detail in the case of molars.

As shown in Table~\ref{tab:real_metrics} and illustrated in Figure~\ref{fig:real_example}, the network has successfully transferred its knowledge from synthetic data to real data, resulting in satisfactory results. We observe that this unified model can complete teeth of various types, e.g., molars, premolars, canines, and incisors. Furthermore, since some teeth were completely removed during the data synthesis process (due to the primitives' volume being too large), the network learned to infer entirely new teeth. Cases \#1 and \#6 show bad metrics, but a visual check shows completed teeth which, however, differ from the ground truth in a notable way. The examples show that there are intradental spaces, and the tops of the teeth seem well aligned, particularly evident in cases \#4 and \#6. In clinical models, the top alignments may differ due to the antagonist impact, as it was present during the modelling. In case \#2, we argue that the fill is even better than in the modelled case, as it seamlessly transitions from the filling to the tooth geometry. However, there are some deficiencies in these examples, particularly the lack of detail in molars \#1, \#3 and premolars \#5. We attribute this to the lower resolution of the SDF representation and to the diffusion process itself.

In Figure~\ref{fig:failure}, we present several failure cases in which the diffusion model did not produce satisfactory results. In case \#7, the network managed to partially generate the tooth, mainly on the side of the neighbouring tooth, but failed on the other side. We attribute this failure to the network's reliance on the neighbouring tooth for context, as it expects tooth 27 in that place. This highlights the network's inability to complete multiple teeth simultaneously, which it was not trained to do. In case \#9, the network generated the upper part of the tooth but failed to connect it to the gingiva, as it became confused by the abutment, which is used to fasten replacements. The \#10 and \#11 cases failed due to unnaturally shaped gingiva. This issue might be resolved by incorporating such examples into the training process. However, the dataset used does not contain examples where the gingiva is extensively modified. All of these factors contribute to the performance drop compared to the results in Section~\ref{sec:modval}
\section{Conclusion}
\label{sec:conclusion}

We present ToothCraft, a unified diffusion-based framework for restoring various classes of partially damaged or missing teeth. Our approach leverages synthetically generated incomplete samples derived from publicly available datasets of segmented dental arches, enabling self-supervised training without the need for curated defect data. Experimental results demonstrate strong performance on synthetic test sets and successful transfer to real clinical scans with missing or damaged teeth.

Despite these promising results, several limitations remain. The model struggles in cases where abnormal gingival anatomy or atypical surrounding context is present, and fine-grained geometric details are limited by the relatively low resolution of the SDF representation due to memory constraints. Addressing these challenges, particularly through higher-resolution representations and more diverse training data, constitutes an important direction for future work.

{
    \small
    \bibliographystyle{ieeenat_fullname}
    \bibliography{main}
}


\end{document}